\documentclass[letterpaper, 10 pt, conference]{ieeeconf}  % Comment this line out if you need a4paper

\IEEEoverridecommandlockouts                              % This command is only needed if 
\usepackage{amsmath} % assumes amsmath package installed
\usepackage{amssymb}  % assumes amsmath package installed
\usepackage{float}
\usepackage{graphicx}

\title{
Strategic Fusion of Vision Language Models: Shapley-Credited Context-Aware Dawid–Skene for Multi-Label Tasks in Autonomous Driving}

\author{Yuxiang Feng $^{1}$, Keyang Zhang $^{1}$, Ashwil Kaniamparambil $^{2}$, Hassane Ouchouid $^{2}$, Ioannis Souflas $^{2}$, \\ and Panagiotis Angeloudis $^{1*}$ % <-this % stops a space
\thanks{*This work was supported by ELM Europe}% <-this % stops a space
\thanks{$^{1}$Yuxiang Feng, Keyang Zhang and Panagiotis Angeloudis are with the Centre for Transport Engineering and Modelling, Department of Civil and Environmental Engineering, Imperial College London, London SW7 2AZ, U.K.}%
\thanks{$^{2}$Hassane Ouchouid, Ioannis Souflas and Ashwil Kaniamparambil are with the ELM Europe, One Canada Square, Canary Wharf, London, E14 5AB, U.K. \newline
$^*$ Corresponding author: Panagiotis Angeloudis (e-mail: p.angeloudis@imperial.ac.uk)}}

\begin{document}

\maketitle
\thispagestyle{empty}
\pagestyle{empty}

%%%%%%%%%%%%%%%%%%%%%%%%%%%%%%%%%%%%%%%%%%%%%%%%%%%%%%%%%%%%%%%%%%%%%%%%%%%%%%%%
\begin{abstract}
Large vision–language models (VLMs) are increasingly used in autonomous-vehicle (AV) stacks, but hallucinations limit their reliability in safety-critical pipelines. We present Shapley-credited Context-Aware Dawid–Skene with Agreement, a game-theoretic fusion method for multi-label understanding of ego-view dashcam video. It learns per-model, per-label, context-conditioned reliabilities from labelled history and, at inference, converts each model’s report into an agreement-guardrailed log-likelihood ratio that is combined with a contextual prior and a public reputation state updated using Shapley-based team credit. The result is calibrated, thresholded posteriors that (i) amplify agreement among reliable models, (ii) preserve uniquely correct single-model signals, and (iii) adapt to drift. To specialise general VLMs, we curate 1{,}000 real-world dashcam clips with structured annotations (scene description, manoeuvre recommendation, rationale) using an automatic pipeline that fuses HDD ground-truth, vehicle kinematics, and YOLOv11 + BoT-SORT tracking, guided by a three-step chain-of-thought prompt; three heterogeneous VLMs are then fine-tuned with LoRA. We evaluate with Hamming distance, Micro-/Macro-F1, and average per-video latency. Empirically, the proposed method achieves a 23\% reduction in Hamming distance, 55\% improvement in Macro-F1, and 47\% improvement in Micro-F1 when comparing with the best single model, demonstrating VLM fusion as a calibrated, interpretable, and robust decision-support mechanism in AV pipelines.
\end{abstract}

%%%%%%%%%%%%%%%%%%%%%%%%%%%%%%%%%%%%%%%%%%%%%%%%%%%%%%%%%%%%%%%%%%%%%%%%%%%%%%%%
\section{INTRODUCTION}
Following GPT-4's release in 2023, large language models (LLMs) have shown strong performance in text processing and are now widely used as general-purpose assistants across professional and technical domains. Extending these capabilities, large vision–language models (VLMs) such as LLaVA \cite{liu2024visual} integrate visual understanding with language, enabling tasks such as image captioning and visual reasoning. These properties are particularly relevant to autonomous vehicles (AVs), where multimodal comprehension, human–machine interaction, and explainability are crucial, and have driven integration of LLMs/VLMs into both modular and end-to-end AV architectures \cite{hanlin_2024}.

In modular pipelines, VLMs enhance scene understanding and contextual reasoning by capturing nuanced interactions, recognising rare or anomalous events, and inferring latent intentions often missed by conventional perception \cite{romero2022optimizing, dewangan2023talk2bev}. Pre-training on large multimodal data supports perception and high-level decision-making, allowing external knowledge, common-sense reasoning, and natural-language instructions to inform planning and improve safety, adaptability, and efficiency \cite{chen2023driving, tanahashi2023evaluation}. For trajectory prediction, VLMs contribute semantic awareness and temporal coherence that aid forecasting of other agents’ motions \cite{li2024driving, peng2024lc}. In end-to-end systems, VLMs have also been used as core components that map raw sensor inputs and textual prompts directly to driving commands, enabling interpretable, goal-directed behaviour across diverse conditions \cite{wen2024road, fu2024drive}.

Despite rapid progress, VLM reliability in safety-critical AV applications remains a subject of debate, primarily due to hallucinations. Modern VLMs can misinterpret complex driving scenes or infer actions from incomplete or ambiguous inputs, producing confident but incorrect outputs. While techniques such as reinforcement learning from human feedback \cite{ouyang2022traininglanguagemodelsfollow} and causal reasoning \cite{li2025treblecounterfactualvlmscausal} have been explored to mitigate hallucinations, studies that directly target VLM-specific hallucinations in the AV context are still limited. As a result, VLMs are typically relegated to auxiliary roles rather than integrated into safety-critical decision pipelines, and their readiness for real-world deployment remains an open research challenge.

To address this challenge, we introduce a game-theoretic fusion method, Shapley-credited Context-Aware Dawid–Skene with Agreement, for multi-label understanding of ego-view dashcam video. Three heterogeneous VLMs each propose a subset of labels from a fixed ontology; from historical labelled rounds we learn per-model, per-label, context-conditioned reliabilities via kernel-weighted Beta-Bernoulli pooling on embedding neighbours, yielding true/false-positive rates as functions of context. During inference, each report is converted to a log-likelihood ratio (LLR), adjusted using pairwise error correlations to guard against double-counting, and weighted by a public reputation state that evolves with Shapley-based team credit. Combined with a contextual prior (from top-K nearest labelled contexts), these signals produce calibrated posterior probabilities per label. This design (i) amplifies agreement among reliable models, (ii) preserves uniquely correct labels proposed by a trusted model, and (iii) adapts over time as performance drifts. Framed as a repeated forecasting game with proper-scoring incentives, the proposed method is interpretable, data-efficient, and yields precise, thresholded probabilities for decision-making.

The main contributions of this study are summarised as follows:
\begin{enumerate}
\item \textbf{Game-theoretic fusion for AV decision-making.} We propose a repeated forecasting framework that operates on ego-view dashcam videos and aggregates three heterogeneous VLMs via context-aware reliabilities (kernel–weighted Beta–Bernoulli on CLIP-embedding neighbours), agreement-weighted log-likelihood ratios with a correlation guardrail, and a contextual prior; calibrated probabilities drive predictions, while Shapley-based team credit and reputation dynamics mitigate hallucinations and adapt trust over time.
\item \textbf{Automatic, temporally coherent annotation pipeline.} We build an annotation system that fuses HDD ground-truth with vehicle kinematics and object-level cues from YOLOv11 + BoT-SORT, orchestrated by a three-step LLaMA-3.2 Chain-of-Thought (scenario description, manoeuvre recommendation, and rationale). This produces structured, time-consistent scene interpretations; 1{,}000 training clips are generated and manually reviewed for quality.
\item \textbf{Task-specialised VLMs and end-to-end evaluation.} Three general-purpose VLMs are adapted to driving via LoRA fine-tuning on the curated data, yielding models that emit manoeuvre recommendations with explicit kinematic cues and interpretable CoT rationales. We report accuracy (Hamming distance, Micro-/Macro-F1) and system performance (average latency) to characterise both effectiveness and efficiency.
\end{enumerate}

\section{RELATED WORK}
A systematic literature review was conducted to assess the current state of the art and shortlist candidate VLMs. A structured Boolean search combined a methods set (“large language model”, “vision–language model”, “vision–language–action model”) with an applications set (“autonomous vehicle”, “self-driving”, “driving instructor”, “robotics”), using OR within sets and AND across them. Searches spanned Scopus, Web of Science, IEEE Xplore, and Google Scholar. 

From an initial pool of results, we shortlisted 42 studies as the most suitable for further analysis. These were classified into three key themes: \textbf{Model}, \textbf{Dataset}, and \textbf{Survey}. Some studies naturally spanned multiple categories. Although the three survey papers provided useful background and suggested evaluation metrics, they were not central to model or dataset development and are therefore not detailed in this study.

\subsection{Model}
Most studies focused on developing or fine-tuning models to understand general or hazard-specific driving scenarios \cite{10654513, takato2024multiframevisionlanguagemodellongform, shi2025scvlmenhancingvisionlanguagemodel, 10495690, 10610797}, and to generate manoeuvre recommendations \cite{chen2023driving, renz2024carllavavisionlanguagemodels, fu2024drive, https://doi.org/10.1002/ail2.56, chen2021reasoninducedvisualattention, 10658060, 10.1007/s10845-023-02294-y, 10607147, 10588851, s24134113, 10561501, LIAO2024100116, 10.1007/978-981-97-5501-1_6, 10549793, Macdonald24, 10.1007/978-3-031-72667-5_15, fan2024navigation, lv2024robomp}. Other papers explored interaction via voice instructions \cite{ryu2024wordswheelsvisionbasedautonomous, HAN202416, 10611525}, or addressed distracted driving detection using in-cabin cameras \cite{hasan2024visionlanguagemodelsidentifydistracted, ZHANG2024107497}. Notably, all reviewed papers were published after 2021, with the majority appearing in 2024.

Many studies utilised Contrastive Language-Image Pre-training (CLIP) or Vision Transformers (ViT) for visual processing. CLIP was frequently selected for its zero-shot generalisation and robust multimodal capabilities \cite{ZHANG2024107497, shi2025scvlmenhancingvisionlanguagemodel, hasan2024visionlanguagemodelsidentifydistracted}. Its visual encoder is typically connected to a projection layer and language generation module for downstream tasks. ViT was also employed in certain studies requiring finer-grained vision tasks \cite{LIAO2024100116, 10611525}, while more traditional approaches such as Mask R-CNN were used for instance segmentation in others \cite{https://doi.org/10.1002/ail2.56}.

For text generation, GPT-series models dominated, appearing in 15 studies, followed by LLaMA and Vicuna in 9. GPT models, developed by OpenAI, are known for strong out-of-the-box performance and API accessibility, but are closed-source. LLaMA, developed by Meta, provides open weights and greater flexibility for research and custom fine-tuning. However, it requires additional instruction-tuning to reach GPT's level of usability in conversational settings. GPT offers immediate utility through commercial services, whereas LLaMA provides a more adaptable platform for bespoke development.

Despite the large number of models, there is no consistent benchmark or head-to-head comparison among them. Many studies do not release pretrained weights or are entirely closed-source, complicating the identification of a definitive state of the art model. A further limitation is that most studies focus on image-text pairs, thereby disregarding the temporal dimension vital for accurate inference of driving event progression.

Nonetheless, two particularly relevant models were identified: HazardVLM \cite{10654513} and VLAAD \cite{10495690}. HazardVLM employs a 3D CNN and GPT-3-style architecture to describe hazards from 7,901 real-world road collision scenes, achieving over 9\% improvement in hazard description accuracy. However, due to challenges replicating its results, it was not selected. VLAAD, on the other hand, uses a pretrained BLIP-2 visual encoder and a frozen LLaMA-2 backbone. It was fine-tuned on three driving datasets and effectively models spatiotemporal relationships within videos, making it a more viable candidate.

To broaden our evaluation scope, more general-purpose VLMs were also considered. Among these, VideoLLaVA \cite{lin2023video} and VideoLLaMA2 \cite{cheng2024videollama}, two leading models on the Microsoft Research Video Description Corpus (MSVD), were shortlisted. VideoLLaVA addresses cross-modal misalignment using a LanguageBind encoder and joint image-video training. VideoLLaMA2 incorporates a spatial-temporal convolution connector to improve video comprehension. Both models demonstrate strong performance across diverse multimodal tasks and offer promising foundations for fine-tuning in this study.

\subsection{Dataset}
Of the 42 reviewed papers, 11 proposed novel datasets tailored for training multimodal models, ranging from several hundred samples \cite{10674252} to over one million \cite{lu2024can}. These datasets generally include both visual and textual data, provided in the form of static images \cite{lu2024can, charoenpitaks2024exploring, 10674252}, video clips \cite{10654513, marcu2024lingoqa, 10446745, 10495690}, or vectorised representations \cite{chen2023driving}.

Hazard-focused datasets such as that used by HazardVLM were deemed too narrow for the broader scope of this study. After evaluating all publicly available datasets, the Honda HRI Driving Dataset (HDD) \cite{ramanishka2018toward} was selected. This dataset contains 104 hours of real-world human driving in the San Francisco Bay Area, segmented into 20-second clips (4,325 for training and 1,525 for testing). Videos are recorded at 1280×720 resolution and 30 fps, and annotated with a four-layer hierarchy: goal-oriented action, stimulus-driven action, cause, and driver attention.

Each clip includes up to five annotated manoeuvres and vehicle state information (e.g., pedal, gear, steering input), providing a rich multimodal source for fine-tuning models. Given its breadth, annotation quality, and relevance, HDD was chosen as the foundation dataset for this study.

\section{METHODOLOGY}
\subsection{Dataset Preparation}
The HDD dataset provides manual annotation for manoeuvres in the format of \textbf{Caption Event} and \textbf{Caption Manoeuvre}. Table \ref{tab:caption} presents an example of video annotations.
\begin{table*}[h!]
    \centering
    \caption{Example of manual annotations in the HDD dataset showing caption events and corresponding manoeuvres} \label{tab:caption}
    \begin{tabular}{p{5cm}|p{9cm}}
    \hline
         \textbf{Caption Event} & \textbf{Caption Manoeuvre} \\\hline
         There is a crosswalk and street parking & drive straight on a road towards a red light, and when it turns green and he continues straight \\\hline
         There is a light and a car in front, & drive straight and then slow down at a red light and then go straight as the light changes green \\\hline
         If there are pedestrians in the crossing, then the driver would stop completely & drive straight until he goes to the light, which is red, but then turns green, so he keeps going \\\hline
         None & proceed down a road in the right lane before stopping at a light behind another car. The light turns green and he proceeds \\\hline
         need to evaluate other drivers while going through the intersection & go straight and stop at red light and start to drive again with green light \\\hline
    \end{tabular}
\end{table*}

While the manual annotations in the selected dataset provide high-level ground-truth for the manoeuvres performed and the associated reasoning, they remain relatively abstract and lack explicit temporal sequencing. This limitation poses challenges when attempting to model the evolving dynamics of a driving scenario or infer causal relationships between observed events and driver actions.

\begin{figure*}[h]
    \centering
    \includegraphics[width=0.95\linewidth]{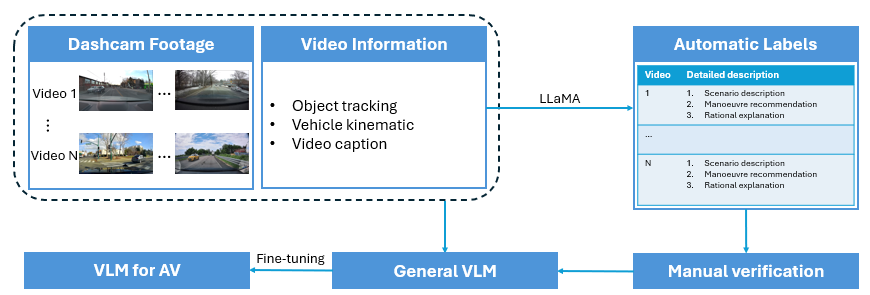}
    \caption{Architecture of the automatic annotation pipeline showing the integration of multiple data sources}
    \label{fig:annotation}
\end{figure*}

An automatic annotation pipeline was developed to reduce dependence on manual annotations. This pipeline integrates ground-truth annotations with vehicle kinematics and object-level information extracted from video frames to generate a more detailed and temporally coherent description of each driving scene. By combining these multimodal inputs, the pipeline can produce enriched scene interpretations that better capture the temporal sequence and interaction of critical events, thereby improving the quality of training data for vision-language-action modelling. The overall architecture of this annotation pipeline is illustrated in Figure \ref{fig:annotation}. 

The annotation pipeline consists of three core components, each contributing to a richer, more structured interpretation of the driving scenario.

The first component involves retrieving ground-truth and vehicle kinematic data from the provided JSON files. The ground-truth annotations offer a high-level understanding of the intended driver behaviour, forming the basis for manoeuvre recommendation. Meanwhile, the kinematic information—such as speed, acceleration, steering angle, and pedal position—enables the identification of the vehicle's current manoeuvre, offering temporal context to the decision-making process.

The second component integrates object tracking results generated using YOLOv11 for object detection and BoT-SORT for multi-object tracking. This module provides detailed spatiotemporal data on both dynamic agents (e.g., pedestrians, vehicles) and static objects (e.g., traffic lights, road signs, barriers) in the scene. The presence or behaviour of these entities often serves as a trigger for changes in the driver's manoeuvres and is thus essential for accurate scenario interpretation.

The third and most crucial component is an LLaMA-3.2 model, guided by a three-step chain-of-thought (CoT) prompting approach. Initially introduced in \cite{wei2022chain}, CoT prompting enhances the reasoning ability of large language models by encouraging step-by-step explanations. Unlike traditional prompting, which produces only a final answer, CoT generates intermediate reasoning steps, yielding more coherent, interpretable outputs across complex tasks. Although initially applied to mathematical reasoning, this method has shown broad applicability in tasks requiring structured logic and justification.

CoT prompting is employed in this pipeline to guide the VLM in producing structured scenario descriptions. The three steps are as follows:
\begin{itemize}
    \item Step 1: Describe the scenario – Provide contextual information such as weather conditions, road layout, and overall driving environment.
\end{itemize}
\begin{itemize}
    \item Step 2: Generate manoeuvre recommendations – Identify and suggest appropriate driving actions in response to the observed scenario. 
\end{itemize}
\begin{itemize}
    \item Step 3: Explain why the manoeuvres are recommended – Explain the rationale behind the suggested manoeuvres, linking them to specific elements of the described scene.
\end{itemize}

This three-step CoT framework is applied throughout the study pipeline, including dataset annotation using LLaMA-3.2, fine-tuning of VLMs for driving-specific tasks, and real-time inference for scenario description and manoeuvre recommendation. CoT ensures that all outputs maintain a consistent, transparent, and human-interpretable structure, which is particularly valuable in safety-critical applications like driver assistance and autonomous planning.

A subset of 1{,}000 videos was randomly selected from the training portion of the dataset to support the fine-tuning process. Each video was processed using the previously described automatic annotation pipeline to generate structured labels. Subsequently, the generated outputs were reviewed through manual inspection to ensure the reliability and accuracy of these annotations.

\subsection{VLM Fine-tuning}
Because the selected models were originally pretrained on general-purpose datasets, their baseline performance when analysing dashcam footage is suboptimal. To realise their full potential and enable the generation of task-specific responses, fine-tuning is required using the prepared domain-relevant dataset.

Fine-tuning is the process of adapting a pretrained machine learning model to a specific application by training it on a smaller, task-specific dataset. Rather than training a model from scratch, which is computationally expensive and data-intensive, fine-tuning leverages the general knowledge acquired during large-scale pre-training and refines it for a specialised use case.

In this study, the fine-tuning process adopts Low-Rank Adaptation (LoRA), a widely used and efficient fine-tuning technique. LoRA injects low-rank update matrices into specific model layers, allowing selective adaptation of key parameters without modifying the entire network. This approach significantly reduces the computational and memory overhead of training while enabling effective specialisation of the model for the target task.

By applying LoRA to the selected vision-language and vision-language-action models, the system can learn driving-specific semantics and scenario dynamics, ultimately improving performance in generating accurate, interpretable manoeuvre recommendations from dashcam inputs.

In conventional fine-tuning methods (e.g., full fine-tuning), the weight matrix \(W\in\mathbb{R}^{d \times k}\) is updated directly, following Equation (\ref{eq:conventional_update}), where \(L\) is the loss function and \(\alpha\) denotes the learning rate. Because all elements of \(W\) are trainable, the total number of trainable parameters for such a fine-tuning method is $\mathcal{O}(dk)$, which is computationally expensive and time-consuming, leading to high fine-tuning costs and low efficiency.
\begin{equation}
W \leftarrow W + \Delta W = W - \alpha \frac{\partial L}{\partial W}
\label{eq:conventional_update}
\end{equation}
Instead of updating the full weight matrix \(W\in\mathbb{R}^{d \times k}\), LoRA keeps the original pretrained parameter matrix \(W_0\in\mathbb{R}^{d \times k}\) frozen, and decomposes \( \Delta W\in\mathbb{R}^{d \times k} \) into two smaller matrices, \( A \) and \( B \), where \( A \in \mathbb{R}^{d \times r} \) and \( B \in \mathbb{R}^{r \times k} \), with the rank \( r \ll \min(d, k) \). \( A \) is initialised with a Gaussian distribution, while \( B \) is initialised to zero. Both \( A \) and \( B \) are learnable during fine-tuning. Finally, the modified weight matrix (i.e., the finetuned parameter matrix) is calculated using Equation (\ref{eq:lora_update}).
\begin{equation}
W = W_0 + \Delta W = W_0 + A B
\label{eq:lora_update}
\end{equation}
LoRA factorises \( \Delta W \) into low-rank matrices \( A \) and \( B \), the total number of parameters being updated is reduced from $\mathcal{O}(dk)$ to $\mathcal{O}(r(d+k))$. As \( r \ll \min(d, k) \), \( \mathcal{O}(r(d+k)) \ll \mathcal{O}(dk) \). The substantial reduction in the number of parameters requiring updates significantly decreases computational overhead and enhances the overall efficiency of the fine-tuning process.

\subsection{Repeated Forecasting Game with Shapley-Based Team Credit}

\paragraph{LLaMA-3.2 Prompting and Label Ontology}
After fine-tuning each VLM on dashcam annotation data, we map free-form outputs to a fixed multi-label ontology using a pretrained LLaMA-3.2 prompt that selects a subset from

\begin{equation}
\mathcal{S}=\left\{
\begin{aligned}
&\text{turn left},\ \text{turn right},\ \text{brake},\ \text{accelerate},\ \text{stop},\\
&\text{traffic light ahead},\ \text{junction ahead},\\
&\text{pedestrian crossing ahead},\ \text{merge},\\ &\text{maintain safe distance},\ \text{check blind spot},\\
&\text{adjust speed due to weather},\ \text{yield to traffic},\\ &\text{drive as normal}
\end{aligned}
\right\}
\end{equation}

yielding a 14 label classification over $\mathcal{L}=\{1,\dots,14\}$. 

\paragraph{Game Protocol and Public State}
Let $w_t=(w_{1,t},w_{2,t},w_{3,t})$ denote the public reputation vector at the start of round $t$. Each round $t=1,\dots,T$ proceeds as follows: 
(i) all models observe the public state $w_t$ and context $x_t\in\mathbb{R}^d$; 
(ii) each model $i\in\{1,2,3\}$ \emph{simultaneously} reports a label set $S_{i,t}\subseteq\mathcal{L}$, which induces per-label binary reports
\begin{equation}
r_{i,t,k}=\mathbb{1}\{k\in S_{i,t}\}\in\{0,1\},\quad i\in\{1,2,3\},\; k\in\mathcal{L}
\end{equation}
(iii) the aggregator computes probabilities $q_t(k)$ using $w_t$, the contextual prior, and the guardrailed signals; 
(iv) ground-truth $y_t(k)\in\{0,1\}$ is revealed (for labelled rounds); 
(v) Shapley-based payoffs $u_{i,t}$ are realised and reputations update to
\begin{equation}
w_{i,t+1}=\frac{w_{i,t}\exp(\eta\,\phi_{i,t})}{\sum_j w_{j,t}\exp(\eta\,\phi_{j,t})}
\end{equation}
(vi) for unlabelled rounds (no $y_t$), we set $u_{i,t}=0$ and $w_{t+1}=w_t$. 
The public state $w_t$ is both prediction-relevant and payoff-relevant, yielding intertemporal incentives. 

\paragraph{Context-aware Reliability}
For each model $i$ and label $k$, we estimate context-conditional reliabilities
\begin{equation}
\theta^+_{i,k}(x)=\Pr(r_{i,t,k}=1\mid y_t(k)=1,\,x_t=x)
\end{equation}
\begin{equation}
\theta^-_{i,k}(x)=\Pr(r_{i,t,k}=1\mid y_t(k)=0,\,x_t=x)
\end{equation}
by kernel-weighted Beta-Bernoulli pooling over a bank $\mathcal{B}$ of past \emph{labelled} rounds
$\{(x_s,r_{i,s,k},y_s(k))\}_{s\in\mathcal{B}}$. With similarity kernel
$\kappa(x,x_s)\!\in[0,1]$ (e.g., cosine$^{\tau}$ on L2-normalised CLIP embeddings $x_t$) and Beta
priors $\alpha_0=\beta_0=1$, the sufficient statistics at query context $x$ are
\begin{align}
\alpha^+_{i,k}(x) &= \alpha_0 + \sum_{s\in\mathcal{N}_K(x)} \kappa(x,x_s)\, r_{i,s,k}\, y_s(k)\\
\beta^+_{i,k}(x)  &= \beta_0  + \sum_{s\in\mathcal{N}_K(x)} \kappa(x,x_s)\, \big(1-r_{i,s,k}\big)\, y_s(k)\\
\alpha^-_{i,k}(x) &= \alpha_0 + \sum_{s\in\mathcal{N}_K(x)} \kappa(x,x_s)\, r_{i,s,k}\, \big(1-y_s(k)\big)\\
\beta^-_{i,k}(x)  &= \beta_0  + \sum_{s\in\mathcal{N}_K(x)} \kappa(x,x_s)\, \big(1-r_{i,s,k}\big)\, \big(1-y_s(k)\big)
\end{align}
where $\mathcal{N}_K(x)$ denotes the top-$K$ nearest neighbors of $x$ (no re-normalization). Posterior means are
\begin{equation}
\hat\theta^+_{i,k}(x)=\frac{\alpha^+_{i,k}(x)}{\alpha^+_{i,k}(x)+\beta^+_{i,k}(x)}
\end{equation}
\begin{equation}
\hat\theta^-_{i,k}(x)=\frac{\alpha^-_{i,k}(x)}{\alpha^-_{i,k}(x)+\beta^-_{i,k}(x)}
\end{equation}
This ensures that in contexts similar to $x_t$, models with historically higher $\hat\theta^+_{i,k}$ and lower $\hat\theta^-_{i,k}$ receive larger LLRs in aggregation, and therefore more weight in the aggregation.

\paragraph{Agreement-weighted Signals and Correlation Guardrail}
Each report contributes the per-model LLR as,
\begin{equation}
\begin{split}
&\lambda_{i,k}(x_t;r_{i,t,k})=r_{i,t,k}\log\frac{\hat\theta^+_{i,k}(x_t)}{\hat\theta^-_{i,k}(x_t)}\\
&+(1-r_{i,t,k})\log\frac{1-\hat\theta^+_{i,k}(x_t)}{1-\hat\theta^-_{i,k}(x_t)}
\end{split}
\end{equation}
To mitigate redundant errors, we compute pairwise error correlations $\rho_{i,j,k}\in[0,\rho_{\max}]$ over a recent labelled window via a smoothed $\phi$-coefficient, and adjust agreeing signals accordingly:
\begin{equation}
\tilde\lambda_{i,k}(x_t;r_{i,t,k})=\frac{\lambda_{i,k}(x_t;r_{i,t,k})}{\,1+\sum_{j\neq i}\rho_{i,j,k}\,\mathbb{1}\{r_{j,t,k}=r_{i,t,k}\}\,}
\end{equation}

\paragraph{Aggregation to Probabilities}
Let $w_{i,t}\ge 0$ with $\sum_i w_{i,t}=1$. Using a contextual prior $\pi_k(x_t)=\Pr(y_t(k)=1\mid x_t)$ calculated from the top-50 nearest labelled contexts, the aggregated probability is
\begin{equation}
q_t(k)=\frac{\pi_k(x_t)\exp\!\Big(\sum_{i} w_{i,t}\,\tilde\lambda_{i,k}(x_t;r_{i,t,k})\Big)}{(1-\pi_k(x_t))+\pi_k(x_t)\exp\!\Big(\sum_{i} w_{i,t}\,\tilde\lambda_{i,k}(x_t;r_{i,t,k})\Big)}
\end{equation}

\paragraph{Coalition Aggregator, Team Score, and Baseline}
For any coalition $C\subseteq\mathcal{N}=\{1,2,3\}$, define the coalition aggregator using logistic sigmoid $\sigma( \mathord{\cdot})$ as,
\begin{equation}
q_t^{(C)}(k)
=\sigma\!\Big(\operatorname{logit}\pi_k(x_t)
+\sum_{i\in C} w_{i,t}\,\tilde{\lambda}^{(C)}_{i,k}(x_t;r_{i,t,k})\Big)
\end{equation}

\begin{equation}
\tilde{\lambda}^{(C)}_{i,k}(x_t;r)
=\frac{\lambda_{i,k}(x_t;r)}{1+\sum_{j\in C\setminus\{i\}}\rho_{i,j,k}\,\mathbb{1}\{r_{j,t,k}=r\}}
\end{equation}

\begin{equation}
v_t(C)=\sum_{k=1}^{K} s_{t,k}\!\big(q_t^{(C)}(k)\big)
-\sum_{k=1}^{K} s_{t,k}\!\big(\pi_k(x_t)\big)
\end{equation}
\begin{equation}
s_{t,k}(q)=y_t(k)\log q + \big(1-y_t(k)\big)\log(1-q)
\end{equation}

\paragraph{Shapley (player-level SHAP) marginal contribution}
With $n=|\mathcal{N}|=3$, the Shapley value for model $i$ in round $t$ is
\begin{equation}
\phi_{i,t}=\sum_{C\subseteq \mathcal{N}\setminus\{i\}}\frac{|C|!\,(n-|C|-1)!}{n!}\,\Big[v_t(C\cup\{i\})-v_t(C)\Big]
\end{equation}
equivalently, label-wise with $s_{t,k}(q)=y_t(k)\log q+(1-y_t(k))\log(1-q)$:
\begin{equation}
\begin{split}
\phi_{i,t}
&=\sum_{k=1}^{K}\sum_{C\subseteq \mathcal{N}\setminus\{i\}}
\frac{|C|!\,(n-|C|-1)!}{n!}\,\Big[ \\
&\quad s_{t,k}\!\big(q_t^{(C\cup\{i\})}(k)\big)-s_{t,k}\!\big(q_t^{(C)}(k)\big)\Big]
\end{split}
\end{equation}

By construction $\sum_i \phi_{i,t}=v_t(\mathcal{N})=S_t(\mathcal{N})-S_t(\varnothing)$.

\paragraph{Stage Payoff and Reputation Dynamics}
Let $P_t\ge 0$ be a team prize and $\alpha\in[0,1]$. The stage payoff and public-state update are
\begin{equation}
u_{i,t}=\phi_{i,t}+\alpha\,P_t\,\frac{w_{i,t}}{\sum_j w_{j,t}}
\end{equation}
\begin{equation}
w_{i,t+1}=\frac{w_{i,t}\exp(\eta\,\phi_{i,t})}{\sum_j w_{j,t}\exp(\eta\,\phi_{j,t})},\quad \eta>0
\end{equation}
Thus present actions affect future influence (via $w_{t+1}$) and future prize shares.

\paragraph{Decision Rule and Calibration}
Predictions are $\hat y_t(k)=\mathbb{1}\{q_t(k)\ge \tau\}$, where $\tau$ is tuned on the training set to maximise the Jaccard index. For numerical stability, all probabilities entering logs are clipped to $[\varepsilon,1-\varepsilon]$ with small $\varepsilon>0$. Initialise $w_{i,1}=1/3$.

\paragraph{Repeated-game Objective}
Each model maximises discounted cumulative utility
\begin{equation}
U_i=\sum_{t=1}^{T}\delta^{\,t-1}u_{i,t},\qquad \delta\in(0,1]
\end{equation}
Stage payoffs are strategically interdependent through $\phi_{i,t}$, and actions change the payoff-relevant public state $w_{t+1}$, establishing a bona fide repeated game. If it is an unlabelled round (\(y_t\) is not observed), we set \(u_{i,t}=0\) and \(w_{t+1}=w_t\). 

\subsection{Evaluation}
We evaluate performance using three metrics: Hamming distance, Micro-F1, and Macro-F1.
% The latency and throughput for video processing is also reported.

\noindent\textbf{Hamming distance (per sample and averaged)}
\begin{equation}
d_{\mathrm{Ham}}(t)
=\frac{1}{K}\sum_{k=1}^{K}\mathbb{1}\{y_t(k)\neq \hat y_t(k)\}
\end{equation}
\begin{equation}
\bar d_{\mathrm{Ham}}
=\frac{1}{N}\sum_{t=1}^{N} d_{\mathrm{Ham}}(t)
\end{equation}

\medskip
\noindent\textbf{Micro-F1 (pooled over labels)}
\begin{equation}
\mathrm{TP}=\sum_{t=1}^{N}\sum_{k=1}^{K}\mathbb{1}\{y_t(k)=1 , \hat y_t(k)=1\}
\end{equation}
\begin{equation}
\mathrm{FP}=\sum_{t=1}^{N}\sum_{k=1}^{K}\mathbb{1}\{y_t(k)=0 , \hat y_t(k)=1\}
\end{equation}
\begin{equation}
\mathrm{FN}=\sum_{t=1}^{N}\sum_{k=1}^{K}\mathbb{1}\{y_t(k)=1 , \hat y_t(k)=0\}
\end{equation}
\begin{equation}
P_{\mathrm{micro}}=\frac{\mathrm{TP}}{\mathrm{TP}+\mathrm{FP}}
\end{equation}
\begin{equation}
R_{\mathrm{micro}}=\frac{\mathrm{TP}}{\mathrm{TP}+\mathrm{FN}}
\end{equation}
\begin{equation}
\mathrm{F1}_{\mathrm{micro}}
= \frac{2 P_{\mathrm{micro}} R_{\mathrm{micro}}}
{ P_{\mathrm{micro}} + R_{\mathrm{micro}} }
\end{equation}

\medskip
\noindent\textbf{Macro-F1 (mean over labels)}
\begin{equation}
\mathrm{TP}_k=\sum_{t=1}^{N}\mathbb{1}\{y_t(k)=1 , \hat y_t(k)=1\}
\end{equation}
\begin{equation}
\mathrm{FP}_k=\sum_{t=1}^{N}\mathbb{1}\{y_t(k)=0 , \hat y_t(k)=1\}
\end{equation}
\begin{equation}
\mathrm{FN}_k=\sum_{t=1}^{N}\mathbb{1}\{y_t(k)=1 , \hat y_t(k)=0\}
\end{equation}
\begin{equation}
P_k=\frac{\mathrm{TP}_k}{\mathrm{TP}_k+\mathrm{FP}_k}
\end{equation}
\begin{equation}
R_k=\frac{\mathrm{TP}_k}{\mathrm{TP}_k+\mathrm{FN}_k}
\end{equation}
\begin{equation}
\mathrm{F1}_{\mathrm{macro}}
= \frac{1}{K} \sum_{k=1}^{K}\frac{2 P_k R_k}{P_k + R_k}
\end{equation}

% \medskip
% \noindent\textbf{Latency (average over $T$ videos)}
% Let $T$ be the number of videos, $s_t$ and $e_t$ denote the start and stop time for processing video $t$, the average latency $\bar T$ can be computed as:

% \begin{equation}
% \bar T = \frac{1}{T}\sum_{t=1}^{T} (e_t - s_t)
% \end{equation}

\section{EXPERIMENT}
The training and evaluation were implemented on a desktop workstation (CPU: Xeon W7-3455@4.8GHz, GPU: 2$\times$NVIDIA GeForce RTX A6000, RAM: 256GB). To streamline deployment, the fusion pipeline is wrapped in a lightweight Flask service that handles routing and request/response logic. The architecture has two core components: (i) an initialisation routine that loads all three VLMs and their weights once at startup, and (ii) an inference routine that processes batches of dashcam videos, runs the models in parallel across available GPUs, and returns text outputs for fusion. Single-load initialisation avoids repeated model setup, while batched, parallel execution improves throughput and ensures efficient processing of multiple videos. 70\% of the annotated videos were used to finetune each VLM and the proposed fusion framework, and 30\% were used for evaluation.

\subsection{Performance Evaluation}
As shown in Table \ref{tab:performance_comparison}, fine-tuned models consistently outperform their pretrained counterparts, and the fusion further advances performance: Micro-F1 = 0.63 and Macro-F1 = 0.31, surpassing all single VLMs. Relative to the best individual baselines, this is a 47\% Micro-F1 improvement over finetuned VideoLLaMA2 (0.43) and a 55\% Macro-F1 improvement over pretrained VideoLLaMA2 (0.20), indicating that the proposed fusion amplifies reliable consensus while preserving uniquely correct signals from individual models. Importantly, the fusion also reduces Hamming distance to 0.17, a 23\% reduction versus the best single model (0.22) and 32\% versus the worst (0.25), reflecting fewer per-instance label mismatches despite multi-label complexity.

The final reputations are $w_{i,T} =[0.34, 0.44, 0.22]$ for VLAAD, VideoLLaMA2, and VideoLLaVA, respectively. This reputation vector implies that more trust is granted to VideoLLaMA2, consistent with its strong per-label balance and complementary evidence; the others remain influential (0.34 and 0.22), indicating non-redundant contributions captured by the Shapley-credited updates. Overall, the fusion provides a more robust and interpretable decision-support signal for AV pipelines.

\begin{table}[h!]
    \centering
    \caption{Model Performance Comparison}
    \begin{tabular}{c|c|c|c}
    \hline
        \textbf{Model} & \textbf{Hamming Distance} & \textbf{Micro-F1} & \textbf{Macro-F1} \\\hline
        \multicolumn{4}{c}{\textbf{Pretrained}} \\\hline
        VLAAD & 0.24 & 0.35 & 0.16 \\\hline
        VideoLLaMA2 & 0.23 & 0.40 & 0.20 \\\hline
        VideoLLaVA & 0.22 & 0.31 & 0.14 \\\hline
        \multicolumn{4}{c}{\textbf{Finetuned}} \\\hline
        VLAAD & 0.23 & 0.40 & 0.19 \\\hline
        VideoLLaMA2 & 0.22 & 0.43 & 0.18 \\\hline
        VideoLLaVA & 0.23 & 0.42 & 0.15 \\\hline
        \multicolumn{4}{c}{\textbf{Fusion}} \\\hline
        \textbf{Majority vote} & \textbf{xxx} & \textbf{xxx} & \textbf{xxx} \\\hline
        \textbf{Proposed model} & \textbf{0.17} & \textbf{0.63} & \textbf{0.31} \\\hline
    \end{tabular}
    \label{tab:performance_comparison}
\end{table}

\subsection{Ablation Study of Model Components}
To evaluate the influence of each component, an ablation study is conducted. First, the reputation of each VLM is frozen, where a uniform $w_i=0.33$ is used to test whether adapting global influence from past marginal contributions $\phi_{i,t}$ is necessary. Second, the Shapley-based team credit is replaced with a simple per-model score without coalition baselines, to probe whether principled attribution of interaction effects, rather than raw accuracy, drives improvements. Third, the correlation guardrail is disabled by setting $\rho_{i,j,k}=0$, eliminating agreement-based shrinkage of signals and revealing the value of mitigating redundant, error-correlated votes. Fourth, the context awareness is removed by replacing the context-conditioned reliabilities $\hat\theta^\pm_{i,k}(x)$ and contextual prior $\pi_k(x)$ with context-agnostic estimates, thereby testing whether per-label, per-scene trust adaptation is required beyond a single global notion of reliability.

\begin{figure}[h]
    \centering
    \includegraphics[width=0.95\linewidth]{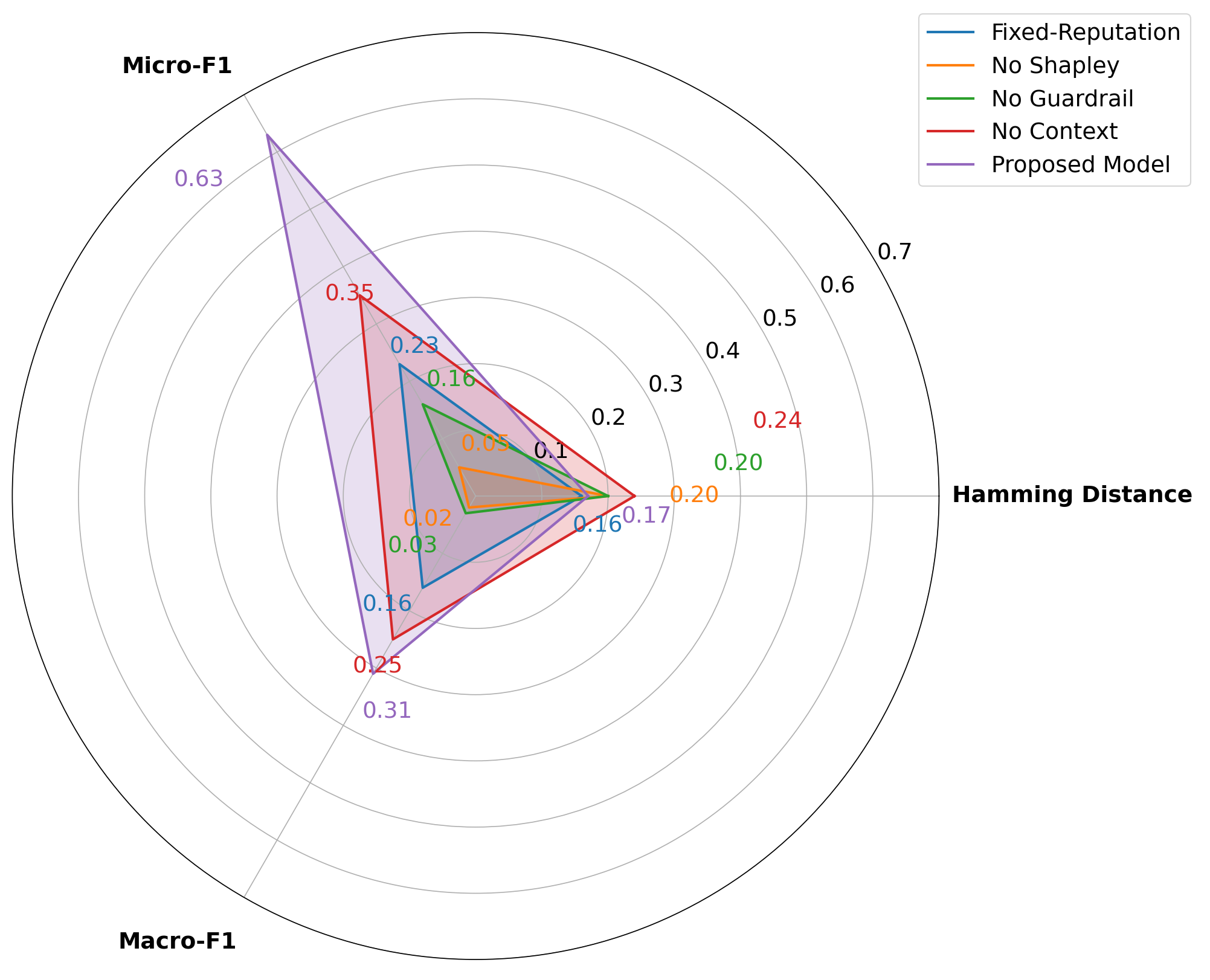}
    \caption{Ablation study of model components}
    \label{fig:ablation}
\end{figure}

As shown in Fig. \ref{fig:ablation}, the ablation analysis indicates that each module contributes a distinct and complementary utility to the fusion framework. Adaptive reputation weighting substantially improves utility by converting consensus into correct positives: fixing $w_i$ yields a marginally lower Hamming distance (0.16) but substantially reduced Micro-/Macro-F1 (0.23/0.16), evidencing the necessity of weight adaptation to historical marginal contributions. Shapley-based team credit is the principal driver of performance—substituting a naive per-model score precipitates a collapse to 0.05/0.02 Micro-/Macro-F1 (Hamming 0.20), demonstrating the importance of principled attribution under interaction effects. The correlation guardrail is essential for calibration; disabling it degrades Micro-/Macro-F1 to 0.16/0.03 and increases Hamming to 0.20, confirming its role in mitigating double-counting from error-correlated agreement. Finally, context awareness delivers per-scene trust that particularly benefits rare labels: removing it improves over naive baselines (0.35/0.25) yet remains well below the full system. Collectively, these components achieve the best trade-off.

\section{CONCLUSIONS}
This study introduces a strategic fusion framework to mitigate hallucinations in autonomous driving by aggregating multiple VLMs. Using the Honda HRI Driving Dataset, we formulate manoeuvre recommendations from ego-view dashcam video as a multi-label task and fine-tune three VLMs (VLAAD, VideoLLaMA2, VideoLLaVA). We then deploy a context-aware Dawid–Skene ensemble with an agreement guardrail and baseline-corrected, Shapley-governed reputation updates, framing the fusion problem as a repeated game. The aggregator adaptively weights model reports by context, consensus, and accumulated performance, producing calibrated, robust label probabilities. Compared with the best individual VLM, our fusion method achieves a 23\% lower Hamming distance, 55\% improvement in Macro-F1, and 47\% improvement in Micro-F1, demonstrating clear gains in both per-instance accuracy and overall precision–recall.

Despite the promising results, our approach has limitations, especially in more challenging settings where model errors are dependent, labels are sparse, or operating conditions shift (e.g., night, rain, unfamiliar roads). The current design assumes largely independent model reports, relies on regular ground-truth for updating reputations and correlations, and is tailored to three models with modest added compute—choices that may constrain performance and real-time deployment in more complex scenarios. Future work will stress-test the method on more challenging and diverse video segments and extended unlabelled sequences to probe robustness. We also plan to scale the credit-and-reputation mechanism to larger ensembles, reduce reliance on labels via active and semi-supervised updates, and strengthen adaptation to distribution shift. Finally, we will examine the trade-offs between richer context modelling and computational efficiency to ensure scalability and real-time applicability across diverse driving conditions.

\bibliographystyle{ieeetr}
\bibliography{ref.bib}

\end{document}